\documentclass[letterpaper, 10 pt, conference]{ieeeconf}  

\IEEEoverridecommandlockouts % This command is only needed if 
\usepackage[utf8]{inputenc}
\usepackage{graphics} 
\usepackage{graphicx} % more modern
\usepackage{mathptmx} 
\usepackage{times} 
\usepackage{amsmath} 
\usepackage{amssymb}
\usepackage{bm}
\usepackage{newtxmath}
\usepackage{newtxtext}
\usepackage{array}
\usepackage{soul}
\usepackage[]{algorithm2e}
\usepackage{xspace}
\usepackage{xcolor}
\usepackage[bottom]{footmisc}

\newcommand{\norm}[1]{\left\lVert#1\right\rVert}

\DeclareMathOperator{\Lhat}{\mathbf{\skew{-5}\hat{L}}}
\DeclareMathOperator{\Hhat}{\mathbf{\skew{0}\hat{H}}}

\newcommand{\acronym}{DeLaN\xspace}

\usepackage{amsmath,amsfonts,bm}

\def\eqref#1{equation~\ref{#1}}
\def\Eqref#1{Equation~\ref{#1}}
\def\1{\bm{1}}

\DeclareMathAlphabet{\mathsfit}{\encodingdefault}{\sfdefault}{m}{sl}
\SetMathAlphabet{\mathsfit}{bold}{\encodingdefault}{\sfdefault}{bx}{n}

\DeclareMathOperator*{\argmin}{arg\,min}

\usepackage{hyperref}
\usepackage{url}

\title{\LARGE \bf
Deep Lagrangian Networks for end-to-end learning of energy-based control for under-actuated systems}

\author{Michael Lutter$^{1}$, Kim Listmann$^{2}$ and Jan Peters$^{1, 3}$% <-this % stops a space
\thanks{This project has received funding from the European Union’s Horizon 2020 research and innovation program under grant agreement No \#640554 (SKILLS4ROBOTS). Furthermore, this research was also supported by grants from ABB, NVIDIA and the NVIDIA DGX Station.}
\thanks{$^{1}$Michael Lutter and Jan Peters are with the Department of Computer Science,
        Technische Universit\"at Darmstadt, 64289 Darmstadt, Germany
        {\tt\small \{lutter, peters\}@ias.tu-darmstadt.de}}%
\thanks{$^{2}$Kim Listmann is with ABB Corporate Research Center Germany, 
Wallstadter Str. 59, 68526 Ladenburg, Germany
        {\tt\small kim.listmann@de.abb.com}}%
\thanks{$^{3}$Jan Peters is with the Max Planck Institute for Intelligent Systems, 
        Spemannstr. 41, 72076 T\"ubingen, Germany
        {\tt\small jan.peters@tuebingen.mpg.de}}%
}

\begin{document}

\maketitle
\thispagestyle{empty}
\pagestyle{empty}

\begin{abstract}
% HELD
% 1) Stating the problem:
Applying Deep Learning to control has a lot of potential for enabling the intelligent
design of robot control laws.
% DRACHENPROBLEM
% 2) Say why it is interesting:
Unfortunately common deep learning approaches to control, such as deep reinforcement learning, 
require an unrealistic amount of interaction with the real system, do not yield any performance guarantees, and do not make good use of extensive insights
from control theory. In particular,
common black-box approaches -- that abandon all insight from control -- 
are not suitable for complex robot systems.

% HERO SLAYS DRAGON
% 3) Say what your solution is / what it achieves:
We propose a deep control approach as a bridge between the solid theoretical foundations
of energy-based control and the flexibility of deep learning. To accomplish 
this goal, we extend Deep Lagrangian Networks (\acronym) to not only adhere to Lagrangian Mechanics but also ensure conservation of energy and passivity of the learned representation. This novel extension is embedded within a energy control law to control under-actuated systems.
% HERO BRINGS HOME THE TREASURE
% 4) Say what follows from your solution:
The resulting \acronym for energy control (\acronym 4EC) is the first model learning approach using generic function approximation that is capable of learning energy control because existing approaches cannot learn the system energies directly. \acronym 4EC exhibits excellent real-time control on the physical Furuta pendulum and learns to swing-up the pendulum while the control law using system identification does not.
\end{abstract}

\section{Introduction}
Control laws are essential to achieve intelligent robots that enable industrial automation, human-robot interaction and locomotion. The common approach is to manually derive the system dynamics, measure the masses, lengths, inertias of the disassembled mechanical system \cite{albu2002regelung} and finally use these equations to engineer a control law for this specific system. Therefore, this engineering approach requires significant effort. In stark contrast, many learning to control approaches, such as Deep Reinforcement Learning \cite{lillicrap2015continuous, schulman2015trust, schulman2017proximal}, try to learn the control law using black-box methods, and hence, do not require any engineering for the specific system. These black-box methods abandon all insights from control and physics, require millions of samples from the physical systems, do not yield any performance guarantees and require extensive reward shaping to the desired solution \cite{mataric1994reward}  or random seeds \cite{henderson2018deep}.  

\begin{figure}[t] \vspace{+0.25cm}
\centering
\includegraphics[width=\columnwidth, angle=0,trim=0mm 5mm 0mm 5mm]{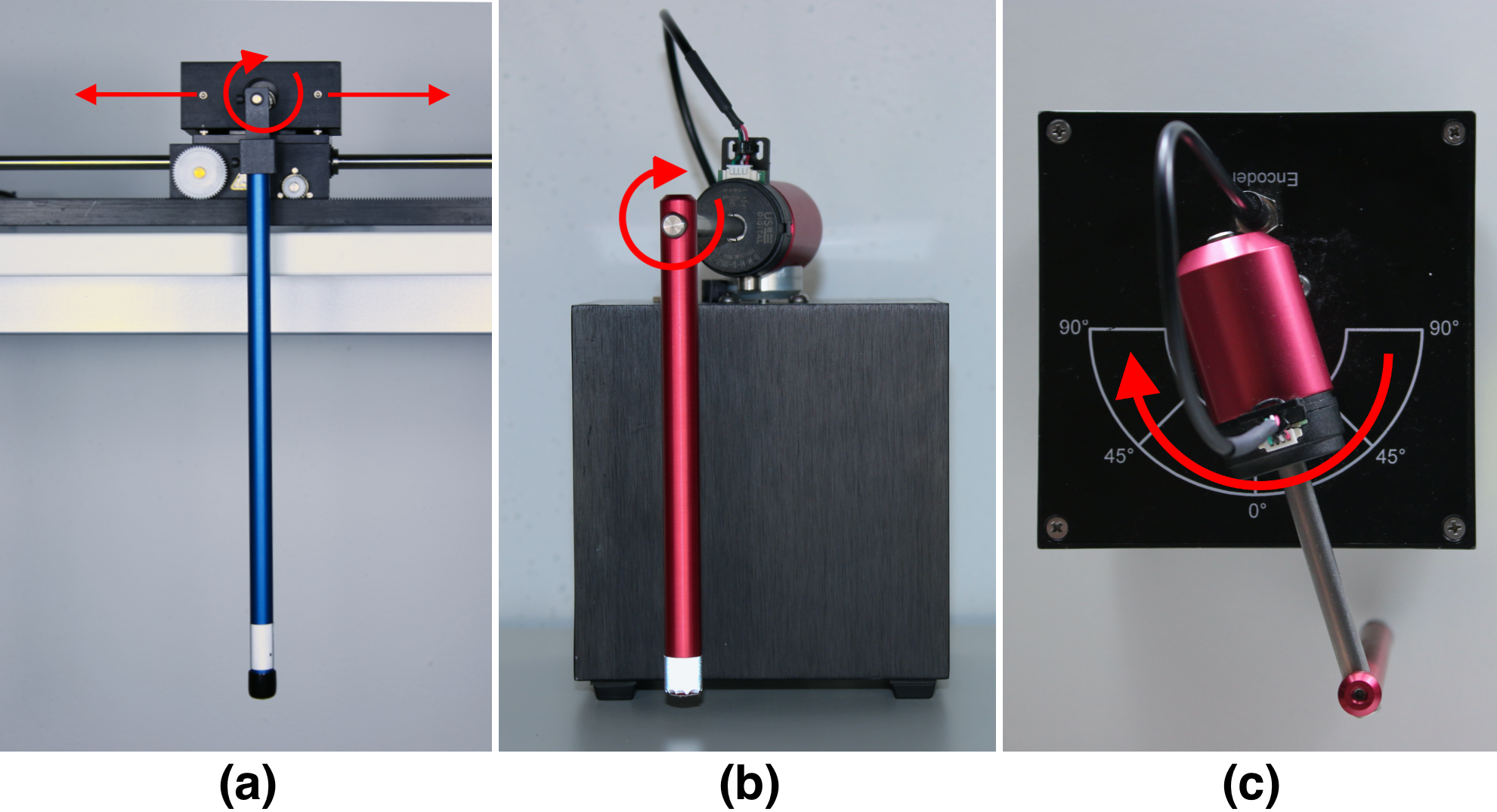}
\caption{The physical Cartpole and Furuta pendulum used to evaluate the energy-based control of under-actuated systems with learned models. (a) The Cartpole consists of a horizontally moving cart driven by a rack and pinion drive. The passive pendulum is attached to the cart and the cart the pendulum can be swung-up and balanced. The front (b) and top (c) view of the Furuta pendulum consisting of an actuated rotary pendulum with a passive vertical pendulum. The vertical pendulum can be swung-up and balanced by moving the horizontal rotary link. Videos for the swing-up using the different models are available at \href{https://youtu.be/m3JRYq7Gmgo}{https://youtu.be/m3JRYq7Gmgo}}
\label{fig:furuta_pendulum}
\vspace{+0.55cm}
\end{figure}

We propose to bridge this gap by combining the flexibility of deep learning with the theoretical insights from control theory in order to achieve learning of control that is independent of the system, applicable for real systems, cannot yield degenerate solutions and requires little engineering. Therefore, we combine existing control laws for energy based control with model learning. Such combination cannot be achieved by standard black-box model learning techniques \cite{schaal2002scalable, choi2007local, nguyen2009model, sanchez2018graph} because these methods learn the mapping from joint state $\{ \mathbf{q}, \: \dot{\mathbf{q}}, \: \ddot{\mathbf{q}} \}$  to motor torques $\bm{\tau}_{M}$, but cannot learn the underlying ODE\footnote{One can also not infer the components of the ODE using the Composite Rigid body algorithm \cite{walker1982efficient} in combination with the learned inverse dynamics mapping, because the black-box function approximation violates the underlying assumptions.} nor the potential and kinetic energies, because these components are not observable and hence, cannot be learned supervised. 
Only our novel extension of Deep Lagrangian Networks (\acronym) \cite{lutter2018deep} is capable of learning the underlying ODE from data using the joint configurations and motor torques. Compared to the previous \acronym, we extend \acronym to also encode energy conservation and coherence besides the Lagrangian Mechanics prior. Therefore, our novel extension of \acronym learns the mass-matrix, the centrifugal, Coriolis, gravitational and frictional forces as well as the potential and kinetic energy using unsupervised learning. Hence, \acronym enables the combination of energy control with learned models without the knowledge of the system kinematics, which must be known for standard system identification techniques \cite{atkeson1986estimation}. In the following we will refer to this combination as \acronym for energy control (\acronym 4EC)

To demonstrate the performance, we apply \acronym 4EC under-actuated systems. This control problem is challenging as the controller must exploit the inherent system dynamics to solve the task and cannot use high-gain feedback control to cancel the system dynamics. For example the swing-up task of the Cartpole and the Furuta pendulum requires the repeated amplification of the amplitude of the passive pendulum before the pendulum can be swung-up. Furthermore, these tasks are a standard evaluation task for learning for control \cite{duan2016benchmarking}. In contrast to most previous research, we apply the control laws also to the physical Furuta pendulum (Figure \ref{fig:furuta_pendulum}) and learn the control without pre-training in simulation.

\subsection*{Contribution}
The contribution of this paper is the novel extension of \acronym and the combination of \acronym and energy control (\acronym 4EC) for controlling under-actuated systems. First, we extend \acronym to incorporate energy conservation and frictional forces. Therefore, the extended \acronym not only adheres to the Lagrangian Mechanics but also ensures energy conservation, temporal coherence of the energy and the passivity of the learned representation. Second, we demonstrate that this combination can achieve energy-control for the Cartpole and the Furuta pendulum. This is demonstrated in simulation and on the physical Furuta pendulum in real-time at $500$Hz and without pre-training in simulation. The performance is compared to the analytic models of the manufacturer as well as the standard system identification approach \cite{atkeson1986estimation}. 

In the following we provide an overview about related work (Section \ref{sec:related_work}), briefly summarize Deep Lagrangian Networks \cite{lutter2018deep} and highlight the proposed extensions to this approach (Section \ref{sec:DeLaN}). Subsequently, we derive our proposed control approach, \acronym for energy control (\acronym 4EC) and state the energy-based control law (Section \ref{sec:DeLaN4EC}). Finally, the experiments in Section \ref{sec:Experiments} evaluate the control performance for simulated and physical under-actuated systems. 

\section{Related Work} \label{sec:related_work}
Controlling under-actuated systems has been addressed from various perspectives including reinforcement learning and control theory. For reinforcement learning the swing-up of passive pendulums is a standard benchmark for continuous state- and action spaces. These methods learn the control policy by treating the control task as black-box and improve the policy using only scalar rewards as feedback. However, most reinforcement learning algorithms can only be used in simulation due to the high sample complexity. Only PILCO \cite{deisenroth2011pilco} learned the Cartpole swing-up on the physical system. 
From a control perspective many control laws for specific under-actuated systems have been proposed \cite{chung1995nonlinear, spong1996energy, fantoni2002non, ishitobi2004swing}. These papers manually derive the dynamics for each system using Lagrangian Mechanics and use the specific equations to derive control laws. For the resulting control laws the stability can be analyzed and guaranteed given the true model \cite{fantoni2002non}. Therefore, the control laws achieve the desired behaviour and cannot exploit ill-posed reward functions but require engineering of the dynamics and control law. 
With \acronym 4EC we use the control perspective and embed a control law within a learning architecture to learn the complete control approach. Rather than using the specific system dynamics for deriving the control law, we use the generic Euler-Lagrange ODE, which describes any mechanical system including closed-loop kinematics, and learn the ODE describing the model from data.   

Learning the model from data has been addressed in the literature by either system identification or supervised black-box function approximation. For system identification the knowledge of the kinematic structure is exploited such that the linkage physics parameters can be inferred using linear regression \cite{atkeson1986estimation}. However, the learned parameters must not necessarily be physically plausible \cite{ting2006bayesian}, can only be linear combinations and can only be applied to kinematic trees \cite{spong2006robot}. In combination with the composite rigid body algorithm \cite{walker1982efficient} the parameters of the Euler-Lagrange ODE including the mass-matrix can be inferred. 
For the function approximations standard machine learning techniques such as Linear Regression \cite{schaal2002scalable, haruno2001mosaic}, Gaussian Mixture Regression \cite{calinon2010learning, khansari2011learning}, Gaussian Process Regression \cite{kocijan2004gaussian, nguyen2009model, nguyen2010using}, Support Vector Regression \cite{choi2007local, ferreira2007simulation}, feedforward- \cite{jansen1994learning, lenz2015deepmpc,ledezma2017first,sanchez2018graph} or recurrent neural networks \cite{rueckert2017learning} have been used. These models learn the forward or inverse mapping from joint configuration $\{\mathbf{q}$, $\dot{\mathbf{q}}$, $\ddot{\mathbf{q}} \}$ to motor torque $\bm{\tau}_{M}$. Therefore, these learned models cannot be used to infer the parameters of the Euler-Lagrange ODE and do not allow a combination with classical control besides inverse dynamics or non-linear feed-forward control \cite{nguyen2010using}. 

In contrast to these existing methods, \acronym learns the Euler-Lagrange ODE directly from data, does not require any knowledge of the kinematic structure and is not restricted to kinematic trees. Therefore, \acronym learns the mass matrix, the centrifugal-, Coriolis-,  gravitational- and frictional forces as well as the system energy using unsupervised learning and fits naturally with  control theory.

\begin{figure*}
\vspace{+0.2cm}
\centering
\includegraphics[width=0.77\textwidth, angle=0,trim=0mm 0mm 0mm 0mm]{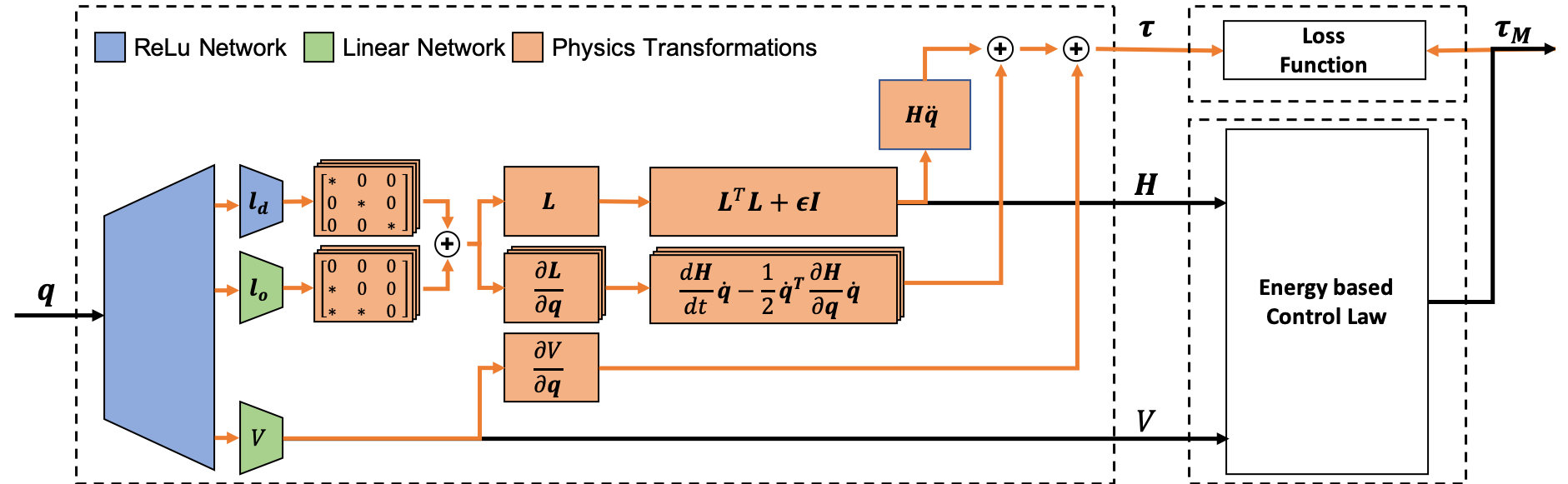}
\caption{The computational graph of the Deep Lagrangian Network for control (\acronym 4EC). Shown in blue and green is the neural network with the three separate heads computing the potential energy $V$ and the mass-matrix $\mathbf{H}$. The orange boxes construct represent the physics transformations constructing Euler-Lagrange equation. For energy-based control these components are directly interfaced to the control law to determine the motor-torque. For training, the gradients are backpropagated through all vertices highlighted in orange.}
\label{fig:ComputationalGraph}
\vspace{-0.2cm}
\end{figure*}

\section{Deep Lagrangian Networks} \label{sec:DeLaN}
First in Section \ref{sec:DLN}, the concept of Deep Lagrangian Networks \cite{lutter2018deep} is summarized and novel extensions are proposed in the subsequent sections. Section \ref{sec:forward} extends the cost function with the forward model. Section \ref{sec:friction} introduces friction such that the Lagrangian Mechanics prior is not violated. Finally, section \ref{sec:energy} adds energy conservation as additional constraint to model learning. Thus, the extended \acronym not only complies with Lagrangian Mechanics but also ensures energy conservation and coherence.  

\subsection{Deep Lagrangian Networks} \label{sec:DLN}
Deep Lagrangian Networks use the knowledge from Lagrangian Mechanics and encode this prior within a deep learning architecture. Therefore, all learned models guarantee that these models must comply with Lagrangian Mechanics. More concretely, let the Lagrangian be defined as $L = T - V$, where $T = 1/2 \: \dot{\mathbf{q}}^{T} \mathbf{H}(\mathbf{q}) \dot{\mathbf{q}} $ is the kinetic energy, $V$ the potential energy and $\mathbf{H}$ the positive definite mass matrix. Substituting $L$ into the Euler-Lagrange differential equation yields the ODE described by
\begin{align} \label{eq:lagrangian_equality}
\mathbf{H}(\mathbf{q}) \ddot{\mathbf{q}} + \underbrace{\dot{\mathbf{H}}(\mathbf{q}) \dot{\mathbf{q}} - \frac{1}{2} \left( \frac{\partial}{\partial \mathbf{q}} \left( \dot{\mathbf{q}}^{T} \mathbf{H}(\mathbf{q}) \dot{\mathbf{q}} \right) \right)^{T}}_{\coloneqq \mathbf{C}(\mathbf{q}, \dot{\mathbf{q}})\dot{\mathbf{q}}} + \: \frac{\partial V}{\partial \mathbf{q}} = \sum_{i} \bm{\tau}_i
\end{align}
where $\bm{\tau}_i$ are the non-conservative generalized forces including motor and frictional forces. Approximating $\mathbf{H}$ and $V$ using deep networks, i.e.,
\begin{align}
    \Hhat = \Lhat(\mathbf{q}; \:\theta) \: \Lhat^{T}(\mathbf{q}; \:\theta) + \epsilon \: \mathbf{I} \hspace{15pt} \hat{V} = \hat{V}(\mathbf{q; \:\psi})
\end{align}
where $\hat{.}$ refers to an approximation, $\Lhat$ is a lower triangular matrix with a non-negative diagonal, $\theta$ and $\psi$ are the network parameters and $\epsilon$ a small positive constant, one can encode the ODE by exploiting the full differentiability of the neural networks \cite{lutter2018deep}. Additionally, the mass matrix $\mathbf{H}$ is guaranteed to be positive definite and the eigenvalues are lower-bounded by $\epsilon$. The network parameters can be learned online and end-to-end, by minimizing the error of the ODE using the samples $\{ \mathbf{q}, \: \dot{\mathbf{q}}, \: \ddot{\mathbf{q}}, \: \bm{\tau}_{M} \}$ recorded on the physical system, i.e. minimizing the $\ell_{i}$ norm between the prediction of \Eqref{eq:lagrangian_equality} and the observed motor torque $\bm{\tau}_{M}$. Therefore, the super-position of the different forces is learned supervised, while the decomposition into inertial, Coriolis, centripetal and gravitational forces is learned unsupervised.

\subsection{Introducing the Forward Model} \label{sec:forward}
Unlike many model learning techniques, \acronym can be used as forward and inverse dynamics model, by solving \Eqref{eq:lagrangian_equality} w.r.t. $\ddot{\mathbf{q}}$. Therefore, one can incorporate the loss of the forward model within the learning of the parameters. This is especially important for many control approaches, including energy control, as these use the inverse of the mass matrix. 
%and the mass matrix usually has high condition number as the links have quite different masses. 
Therefore, incorporating the forward model within the learning of the parameters should yield better approximation of the inverse. Solving \Eqref{eq:lagrangian_equality} for $\ddot{\mathbf{q}}$ yields
\begin{align*}
    % f\left(\mathbf{q}, \: \dot{\mathbf{q}}, \: \bm{\tau}_{M} \right) = 
    \mathbf{H}^{-1}(\mathbf{q}) \left( \sum_{i} \bm{\tau}_i - \dot{\mathbf{H}}(\mathbf{q}) \dot{\mathbf{q}} + \frac{1}{2} \left( \frac{\partial}{\partial \mathbf{q}} \left( \dot{\mathbf{q}}^{T} \mathbf{H}(\mathbf{q}) \dot{\mathbf{q}} \right) \right)^{T} - \: \frac{\partial V}{\partial \mathbf{q}} \right) =  \ddot{\mathbf{q}}.
\end{align*}
Thus the loss function can be extended to minimize the error of the inverse and forward model, i.e.,
\begin{equation}
\begin{aligned} \label{eq:for_loss}
    (\theta^{*}, \psi^{*}) = \argmin_{\theta, \psi} \:\: &\ell_{i}\left(\hat{f}(\theta, \: \psi), \:\: \ddot{\mathbf{q}}\right) + \\ \vspace{-0.00cm} &\ell_{i}\left(\hat{f}^{-1}(\theta, \: \psi), \:\: \bm{\tau}_{M}\right) + \lambda \Omega(\theta, \: \psi)
\end{aligned}
\end{equation}
where $\Omega$ is the $l_2$ weight regularization. 

\subsection{Introducing Friction to Model Learning} \label{sec:friction}
Incorporating friction within model learning in a non black-box fashion is non-trivial because friction is an abstraction to combine various physical effects. For robot arms in free space the friction of the motors dominates, for mechanical systems dragging along a surface the friction at surface dominates while for legged locomotion the friction between the feet and floor dominates but also varies with time. Therefore, defining a general case for all types of friction in compliance with the Lagrangian Mechanics is challenging. Various approaches to incorporate friction models can be found in \cite{lurie2013analytical, wells1967schaum}. Furthermore, if the friction model includes stiction the dynamics are not invertible because multiple motor-torques can generate the same joint acceleration \cite{ratliff2016doomed}. 

This paper focuses on friction caused by the actuators. For actuator friction different models have been proposed \cite{olsson1998friction, albu2002regelung, bona2005friction, wahrburg2018motor}. These models assume that the motor friction only depends on the joint velocity $\dot{\mathbf{q}}_i$ of the $i$th-joint and is independent of the other joints \cite{olsson1998friction, albu2002regelung, bona2005friction, wahrburg2018motor}. Depending on model complexity a combination of static, viscous or Stribeck friction is assumed as model prior and the superposition is described by 
\begin{align} \label{eq:friction_prior}
\bm{\tau}_{f_i} = - \left( \tau_{C_{v}} + \tau_{C_{s}} \: \exp\left(-\dot{\mathbf{q}}_i^{2} / \nu  \right) \right) \text{sign}\left(\dot{\mathbf{q}}_i \right) - d  \: \dot{\mathbf{q}}_i
\end{align}
where $\tau_{C_{v}}$ is the coefficient of static friction, $d$ the coefficient of viscous friction, and $\tau_{C_{s}}$ and $\nu$ are the coefficients of Stribeck friction. In the following the friction coefficients are abbreviated as $\phi = \{\tau_{C_{v}}, \: \tau_{C_{s}}, \: \nu, \: d \}$. Since the frictional force $\bm{\tau}_{f}$ is a function of the generalized coordinates, the frictional force is a non-conservative and generalized force and can simply be added to the Lagrange Euler ODE (\Eqref{eq:lagrangian_equality}). For other types of friction this is not true and one needs to explicitly ensure that one can express the frictional force as generalized force. Given the model prior of \Eqref{eq:friction_prior} the friction coefficients $\phi$ can be learned by treating the coefficients as network weights.

\begin{figure*}
\vspace{+0.1cm}
\centering
\includegraphics[width=\textwidth, trim=0mm 6mm 0mm 0mm]{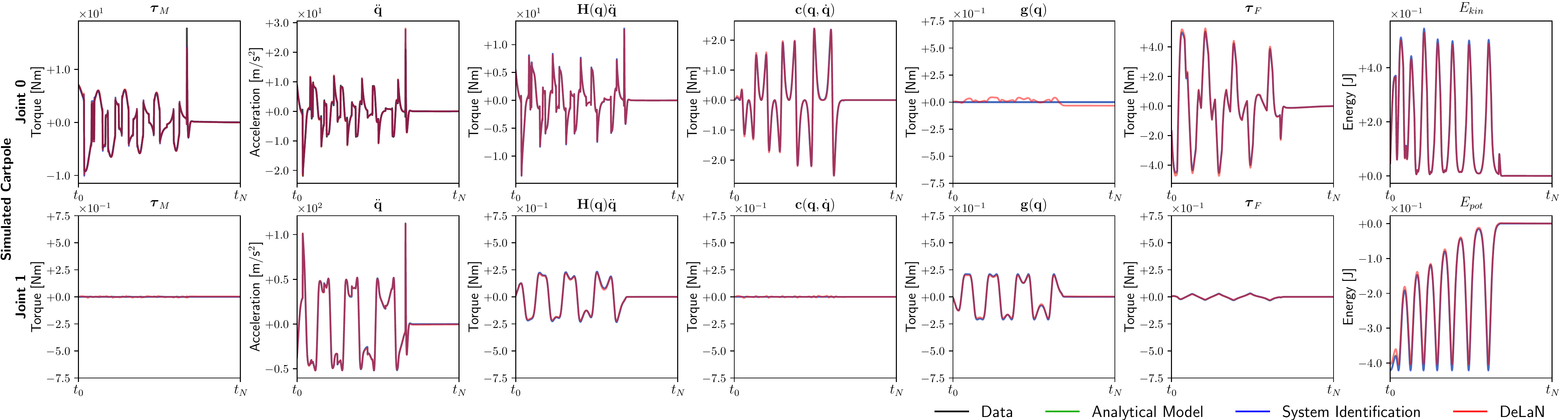}
\includegraphics[width=\textwidth]{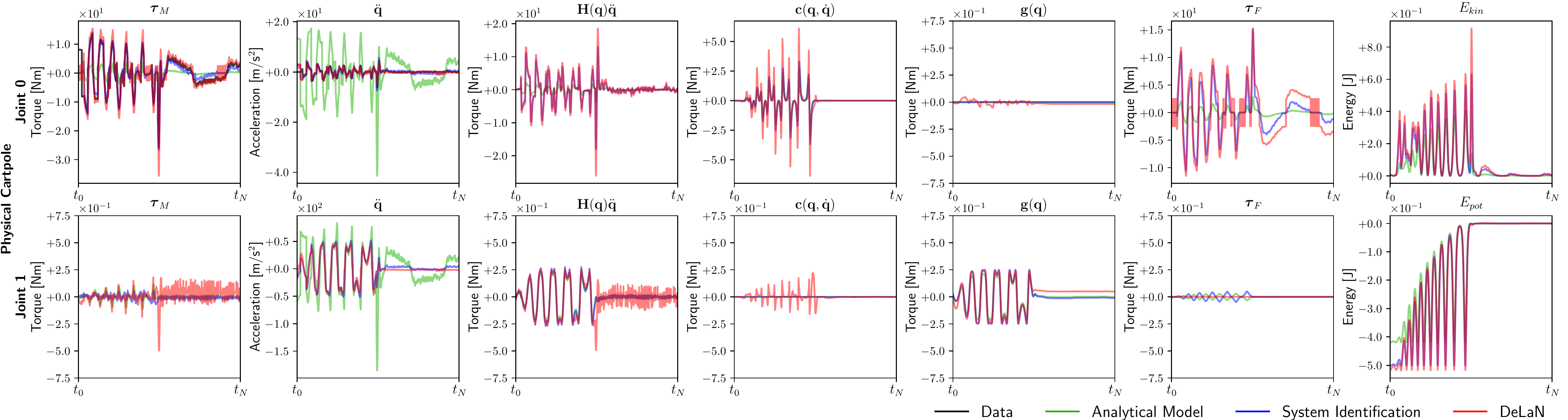}
\caption{The motor-, centrifugal-, Coriolis-, gravitational- and frictional forces, the joint acceleration as well as the kinetic and potential energy for the swing-up of the simulated and physical Cartpole. Using only the super-imposed motor torques and joint accelerations as supervising feedback, \acronym learns to disambiguate between the individual force components and system energies.}
\label{fig:decomposition}
\vspace{-1em}
\end{figure*}

\subsection{Introducing Energy to Model Learning} \label{sec:energy}
Besides the Lagrangian Mechanics objective, incorporating energy conservation and energy coherence, i.e., ensuring that $E_i(t) \:\: \forall t \geq 0$ is at least of Class $C^{2}$, is natural because the Lagrangian $L$ contains the system energy. In order to ensure the conservation of energy,  the total energy of the system must be equal to the summation of the initial system energy $E_0$, the work done by the actuators $W_m$ and the energy losses due to friction $E_{th}$, i.e., 
\begin{align} \label{eq:energy}
E(t) = T(t) + V(t) = E_{0} + W_{M}(t) + E_{th}(t) \hspace{15pt} \forall t \geq 0.
\end{align}
The actuator work or the losses to friction can be computed by numerical integration described by
\begin{align}
    W_{j}(t) = \int^{\mathbf{q}(t)}_{\mathbf{q}(0)} \bm{\tau}^{T}_{j}(\mathbf{q}) \:\: d\mathbf{q} = \int^{t}_{0} \bm{\tau}^{T}_{j}(\mathbf{q}(u)) \:\: \dot{\mathbf{q}}(u) \:\: du
\end{align}
where $\bm{\tau}_{j}$ is either the frictional torque $\bm{\tau}_{F}$ or actuator torque $\bm{\tau}_{M}$ and $W_{j}(0) \coloneqq 0$. This can also be expressed in using the change in energy, i.e., 
\begin{equation} \label{eq:power}
\begin{aligned} 
    \dot{E} = \dot{\mathbf{q}}^{T} \left(\bm{\tau}_{M} + \bm{\tau}_{F} \right) &= \dot{T} + \dot{V} \\
    &= \dot{\mathbf{q}}^{T} \: \mathbf{H} \: \ddot{\mathbf{q}} + \frac{1}{2} \dot{\mathbf{q}}^{T} \: \dot{\mathbf{H}} \: \dot{\mathbf{q}} + \dot{\mathbf{q}}^{T} \: \frac{\partial V}{\partial \mathbf{q}}.
\end{aligned}
\end{equation}
Following \cite{spong2006robot} and recognizing that $\bm{\tau}_{M} + \bm{\tau}_{F}$ is the total force acting on the mechanical system, \Eqref{eq:power} not only ensures energy conservation but also the passivity of the learned system, because this equality ensures the lower bound on the total energy, i.e., $E(T) - E(0) \geq -E(0) \:\:\: \forall \:\: T > 0$. Therefore, the learned model representation is guaranteed to be passive on the training domain given sufficiently low training error. This property of \acronym implies that the uncontrolled system described by the learned dynamics is stable. For black-box function approximation methods this must not be necessarily be true because these methods can learn an active system that is optimal w.r.t. the given cost function. Besides the conservation of energy, the energy coherence can be used as additional constraint ensuring that both the kinetic and potential energy is continuous and differentiable w.r.t. time, i.e., $T, V \in C^{1}$. Using a first order Taylor approximation this constraint can be expressed as
\begin{align} \label{eq:kinetic_coherence}
    \tilde{T}(\mathbf{q}_{t+\delta t}; \: \theta) &= 
    %\int^{t}_{0} \: \dot{\mathbf{q}}^{T}  \frac{\partial \hat{T}(\mathbf{q}, \, \dot{\mathbf{q}}; \: \theta)}{\partial \mathbf{q}} + \: \ddot{\mathbf{q}}^{T} \: \frac{\partial \hat{T}(\mathbf{q}, \, \dot{\mathbf{q}}; \: \theta)}{\partial \dot{\mathbf{q}}} \:\:dt = 
    \hat{T}(\mathbf{q}_{t}; \: \psi) + \dot{\mathbf{q}}^{T}_{t} \: \mathbf{H} \: \ddot{\mathbf{q}}_{t} \delta_t + \frac{1}{2} \dot{\mathbf{q}}^{T}_{t} \: \dot{\mathbf{H}} \: \dot{\mathbf{q}}^{T}_{t} \: \delta_t
    \\
    \label{eq:potential_coherence}
    \tilde{V}(\mathbf{q}_{t+\delta t}; \: \psi) &= 
    %\int^{t}_{0} \dot{\mathbf{q}}^{T} \:\: \frac{\partial \hat{V}(\mathbf{q}; \: \psi)}{\partial \mathbf{q}} \:\:dt =
    \hat{V}(\mathbf{q}_{t}; \: \psi) + \dot{\mathbf{q}}^{T}_{t} \: \frac{\partial \hat{V}}{\partial \mathbf{q}} \: \delta_t.
\end{align}
The resulting equations cannot be directly used as a loss because the true kinetic- and potential energy of the configuration $\mathbf{q}_{t+\delta_t}$ is unknown. Therefore, we bootstrap the current approximation of $\tilde{V}$ and $\tilde{T}$ as target value and do not propagate the gradients through these estimates. In addition, the energy for a specific joint configuration $\mathbf{q}^{*}$ is clamped to a pre-specified value as in \cite{riedmiller2005neural}, i.e. $E(\mathbf{q}^{*}, \: \dot{\mathbf{q}}^{*}) \coloneqq 0$. %
Adding energy conservation (\Eqref{eq:power}) and energy coherence (\Eqref{eq:kinetic_coherence} \& \Eqref{eq:potential_coherence}) to the optimization problem of \Eqref{eq:for_loss} yields the loss for \acronym 4EC.

\section{Deep Lagrangian Networks for Energy Control} \label{sec:DeLaN4EC}
In the previous section, we showed that \acronym can learn the mass matrix, the centripetal, gravitational and frictional forces as well as the kinetic and potential energies using only the joint measurements $\left( \mathbf{q}, \: \dot{\mathbf{q}}, \: \ddot{\mathbf{q}} \right)$ and the actuator torques $\bm{\tau}_{M}$. Using these properties, energy control can be achieved by embedding the learned energies within a energy-based control law. Therefore, \acronym 4EC enables the control of a large-class of under-actuated systems, because these systems are mainly controlled using energy-based control laws and other black-box identification techniques cannot learn the system energy and hence, cannot be applied .   
For energy control, the control law proposed by Spong et. al. \cite{spong1996energy}, which is applicable to the Furuta pendulum, the Cartpole and the Acrobot is used. This control law regulates the energy of the pendulum $E_{p}$ to obtain the desired energy $E^{*}$ and adds an additional P-controller on the active joints to avoid the joint limits. For systems with high friction an additional term to compensate the friction of the actuator can be added. Expressing this control law using the mass-matrix and the potential energy is described by 
\begin{align} \label{eq:u_E}
    \mathbf{u}_{E} &= k_E \left(E_{p} - E^{*} \right) \: \text{sign}\left( \dot{\mathbf{q}}_{p} \cos(\mathbf{q}_p \right) + \mathbf{K}_p \left( \mathbf{q}^{*}_a - \mathbf{q}_a \right)% \\
    %&\hspace{45pt}E_{p} = \frac{1}{8}  \mathbf{q}^{T}_p \:  \mathbf{H}_{22} \: \mathbf{q}_p + V(\mathbf{q}) \nonumber.
\end{align}
with the pendulum energy $E_{p} = 1/8 \:\: \dot{\mathbf{q}}^{T}_p \:  \mathbf{H}_{22} \: \dot{\mathbf{q}}_p + V(\mathbf{q})$ and the desired energy $E^{*}(\mathbf{q}^{*}, \dot{\mathbf{q}}^*)$ at the desired joint configuration~$\mathbf{q}^*$.  

\begin{figure}[t]
\centering
\includegraphics[width=\columnwidth, trim=0mm 0mm 0mm 0mm]{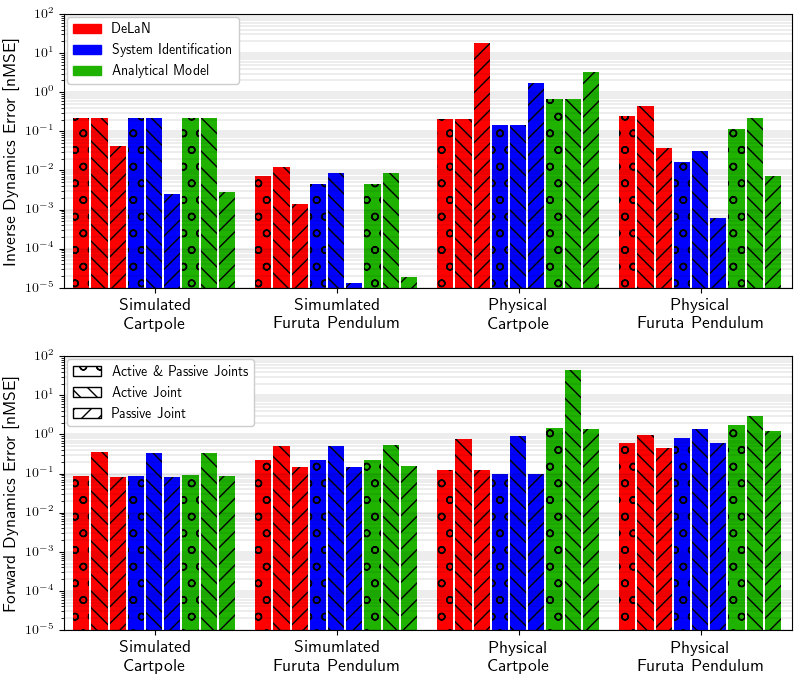}
\caption{The normalized mean squared error of the forward and inverse models for the simulated and physical platforms on the test set.}
\label{fig:nmse}
\vspace{-2em}
\end{figure}

\begin{figure*}
\vspace{+0.1cm}
\centering
\includegraphics[width=\textwidth, angle=0,trim=0mm 0mm 0mm 0mm]{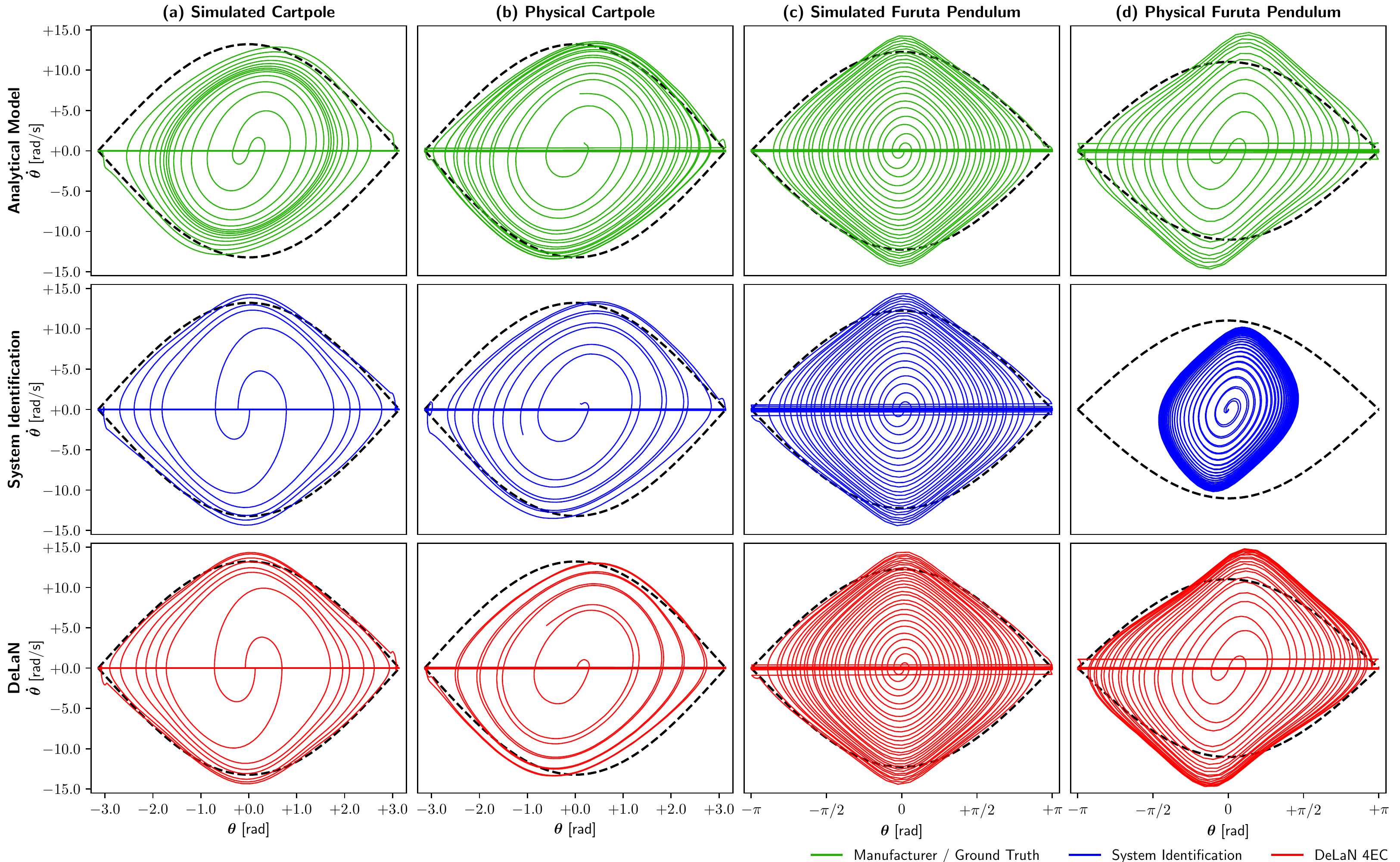}
\caption{The position $\theta$ and velocity $\dot{\theta}$ orbits recorded using energy control to swing-up the passive pendulums. The rows show the different models, i.e., the analytic model, the system identification model and the \acronym model while the columns show the different simulated and physical systems. The dahsed orbit highlights the desired energy $E^{*}$. While the learned and the analytic model can swing-up the simulated system and physical Cartpole only the analytic model and \acronym 4EC can swing-up the physical Furuta pendulum, while the energy controller using the System Identification model cannot.}
\label{fig:orbits}
\end{figure*}

\section{Experiments} \label{sec:Experiments}
We apply \acronym 4EC to control two under-actuated systems: the Cartpole (Figure \ref{fig:furuta_pendulum}a) and the Furuta pendulum (Figure \ref{fig:furuta_pendulum}b). The Cartpole is a horizontally moving cart with an attached passive pendulum. Moving the cart horizontally indirectly controls the pendulum and using this indirect control the pendulum can be swung-up and balanced. Similarly, the Furuta pendulum (also referred to as whirling pendulum) consists of an actuated rotary pendulum with a vertical passive pendulum. Using the rotary pendulum the vertical link can be swung-up and balanced. These experiments are standard experiments for learning to control. However, most previous research only used simulations while we apply these methods to the physical Cartpole and Furuta pendulum.

\subsection{Experimental Setup}
To learn the control task, a smooth exploration policy, i.e., the energy controller using the analytic model, interacts for $T$ seconds with the system and generates data containing $\{ \mathbf{q}, \: \dot{\mathbf{q}}, \:  \ddot{\mathbf{q}}, \: \bm{\tau}_{M}\}_{1 \dots N}$. For the Furuta pendulum the interaction time is $120$s while the interaction time for the Cartpole is $240$s. Using these highly correlated samples the control is learned offline. After learning the controller, the performance is evaluated using the normalized mean square error (nMSE) on the test data (Section \ref{sec:offline}) as well as the online control performance on the swing-up tasks (Section \ref{sec:control}). The online control evaluation is the more relevant performance measure as the nMSE can be deceiving and a low nMSE does not necessarily imply a good control performance. For the swing-up task the systems are first stabilized to the desired energy $E^{*}$ of the balancing point using energy control and then balanced at the unstable equilibrium using a PD-controller. Both, controller operate at $500$Hz and the gains are tuned for each system using the analytic model provided by the manufacturer and fixed afterwards for each experiment. 

The simulated experiments are performed using Bullet \cite{coumans2018} with joint torque as control input. The physical experiments are performed using the \href{https://www.quanser.com/products/linear-servo-base-unit-inverted-pendulum/}{Cartpole} (Fig. \ref{fig:furuta_pendulum}a) and \href{https://www.quanser.com/products/qube-servo-2/}{Furuta pendulum} (Fig. \ref{fig:furuta_pendulum}b) manufactured by Quanser. These physical systems are directly controlled using the DC motor voltage. For the experiments the voltage to motor-torque conversion is performed using the parameters of the manufacturer. Furthermore, both physical systems have unique properties that make the model learning and the control challenging. The linear actuation of the Cartpole is a pinion \& rack drive causing significant stiction and this stiction renders the model learning challenging. In contrast, the links of the Furuta pendulum are very light weight and even small errors of motor voltages push the active joints to its joint limit and stop the episode.

% \subsection{Baselines}\vspace{-0.2cm}
The performance of \acronym 4EC is compared to the dynamics parameters of the manufacturer and the white-box system identification introduced by \cite{atkeson1986estimation} with the extension of viscous friction as in \cite{ting2006bayesian}. For system identification the mass matrix is computed using the Composite Rigid Body algorithm \cite{walker1982efficient} and the potential energy is computed using the analytic expression $V(\mathbf{q}) = m g l \hspace{2pt}(\cos(\mathbf{q}) + 1)$, where only the mass $m$ is inferred from data while the gravitational constant $g$ and pendulum length $l$ are pre-defined constants. These requirements are in stark contrast to \acronym 4EC, because the white-box system identification approach requires the kinematic structure defining the link length, connection between links and gravitational constant, while \acronym 4EC must learn the kinematic and dynamic structure from data. Furthermore, the assumption of knowing the kinematic structure simplifies the learning of the potential energy to merely fitting the amplitude of the potential energy while \acronym 4EC must not only learn the amplitude but also learn the shape. We do not compare to other black-box learning techniques such as neural networks or Gaussian process regression because these techniques cannot learn energy and hence, cannot be applied to energy control.

\subsection{Offline Evaluation} \label{sec:offline}
Figure \ref{fig:decomposition} shows the learned dynamics model and energies of the simulated and physical Cartpole executing a swing-up using the control law described by $\mathbf{u}_{E}$. The dynamics learned by \acronym closely resemble the data as well as the frictional-, inertial-, centrifugal-, Coriolis- and gravitational forces predicted by the analytic model. Furthermore, the kinetic and potential energies are learned. Only the learned potential and kinetic energy for the physical system are slightly scaled similar to the system identification model. Although the energies are scaled, both the kinetic and potential energy are scaled coherently such that the energy conservation holds as enforced by \Eqref{eq:potential_coherence}. Furthermore, \acronym learns the high stiction $\bm{\tau}_F$ of the pinion and rack drive and predicts the close to zero accelerations $\ddot{\mathbf{q}}$ and non-zero motor torques $\bm{\tau}_M$ during balancing of the physical Cartpole, given a sufficiently good initialization of the friction model. The analytic model and system identification cannot represent the stiction and predict either zero torques and non-zero accelerations or zero torques and zero acceleration. Both predictions oppose the measured data. However, \acronym suffers from high frequency noise on the passive pendulum or close to zero components, whereas the white box models do not because the white box models consist of global parameters, which are not susceptible to noise and these models can exploit the known kinematic structure to infer zero Coriolis or gravitational forces. 

Figure \ref{fig:nmse} shows the quantitative comparison using the nMSE defined as
\begin{align}
    \text{nMSE} = \frac{\sum_{i=0}^{N} \norm{\mathbf{x}_i - \hat{\mathbf{x}}_i}_2^2}{\sum_{i=0}^{N} \norm{\mathbf{x}_i + \delta}^2_2} 
\end{align} 
whereas $\delta$ is a small constant for numerical stability. The nMSE is evaluated on test data performing a swing-up, which in the case of the Cartpole is identical to the data shown in Figure \ref{fig:decomposition}. For the simulations the analytic model is the true model, which has a non-zero nMSE because noise is added to the torques during simulation and the accelerations are computed using finite differences and are low-pass filtered because this signal-processing is required for the physical system. The comparison shows that \acronym obtains a similar nMSE as system identification and the analytic model for the forward model of the simulated systems. For the inverse model of the simulated systems, \acronym obtains comparable nMSE for the actuated joint but slightly increased nMSE for the passive joints because \acronym is susceptible to noise and the nMSE is very sensitive to the noise of the passive joint as $\bm{\tau}_p \coloneqq 0$. For the physical systems, \acronym and system identification obtain a lower nMSE than the analytic model for the forward model. For the inverse model, both learned models achieve better performance than the analytic model on the active joint and only for the passive joint \acronym performs slightly worse due to the noise. This noise is negligible during optimization and control because the MSE is dominated by the actuated joint and only the nMSE per joint amplifies the impact of the noise. Overall, the qualitative and quantitative evaluation showed that the performance of the learned models in comparison to the analytic model achieve comparable performance for the simulated systems and a slightly better performance for the physical systems.

\begin{table}[h]
  \centering
  \caption{Percentage of successful swing-ups of simulated and physical Cartpole and Furuta pendulum for the different models.}
  \label{tab:performance}
  \begin{tabular}{|p{0.35\columnwidth}|c|c|c|c|} 
    \hline
    & \multicolumn{2}{c|}{\textbf{Cartpole}} & \multicolumn{2}{c|}{\textbf{Furuta}} \\
    \hline
     \textbf{Model} & \multicolumn{1}{c|}{Sim} & \multicolumn{1}{c|}{Real} & \multicolumn{1}{c|}{Sim} & \multicolumn{1}{c|}{Real} \\
    \cline{2-5}
    %\hhline{~--}
    \hline
    Analytic Model          & 1.00 & 1.00 &  1.00 &  1.00    \\ \hline
    System Identification   & 1.00 &  1.00  & 0.93  &  0.00  \\ \hline
    \acronym                & 1.00 &  1.00  & 1.00  & 0.90     \\ \hline
  \end{tabular}
\end{table}

\subsection{Online Control Evaluation} \label{sec:control}
For the online control experiments the energy-based control law described in \Eqref{eq:u_E} is applied with a control frequency of $500$Hz to 30 different initial joint configurations. Therefore, all models must achieve real-time computation of at least $500$Hz on the physical system to be able to solve the task. For the simulated experiments, the starting configuration is randomly sampled while the physical experiments are performed sequentially and hence, the starting configuration naturally changes. For the physical Cartpole we augment the energy controller with an negative derivative gain to compensate the large viscous friction of the pinion and rack drive. The percentage of successful swing-ups is summarized in Table \ref{tab:performance} and the corresponding position $\theta$ and velocity $\dot{\theta}$ orbits for two starting configurations of the passive pendulum are shown in Figure \ref{fig:orbits}.  Videos of the swing-up of the physical Cartpole and physical Furuta pendulum can be found at \href{https://youtu.be/m3JRYq7Gmgo}{https://youtu.be/m3JRYq7Gmgo}.  

For the simulated systems the analytic and the learned models achieve the successful completion of the swing-up. Only the system identification model fails on 2 trials. These unsuccessful completions are caused by the balancing controller because the system identification model swings-up the pendulum with slightly too much or too low energy such that the PD-controller fails to stabilize the pendulum. Furthermore, the resulting trajectories for the learned models are indistinguishable. For the physical Cartpole all models achieve smooth real-time control and swing-up the pendulum for 30 consecutive times from varying starting configurations.    
For the physical Furuta pendulum only the analytic model and \acronym 4EC achieve the successful swing-up, while the system identification model can only stabilize the pendulum to a low amplitude cycle, which does not reach the balancing point. For 30 trials, \acronym 4EC achieves the successful completion for 27 trials. The three unsuccessful trials are caused by the PD-Controller not being able to stabilize the pendulum because \acronym 4EC swings up the pendulum with slightly less energy than the analytic model and the balancing PD-Controller is very sensitive to these changes in velocity at the switching point. Tuning the PD-controller gains w.r.t. to \acronym 4EC results in the successful completion of all 30 trials but decreases the performance of the analytic model. For fair comparison the gains were fixed between the experiments and optimized for the analytic model. The system identification model fails to swing-up the pendulum because this approach learns a too low mass for the pendulum and hence, can only stabilize the pendulum to a low amplitude oscillation. The learning fails because the regressor of the system identification has too low rank and can only infer a linear combination of the dynamics parameters \cite{siciliano2016springer}.

\acronym 4EC is capable of solving the swing-up for the simulated and physical Cartpole and Furuta pendulum using a 500Hz real-time control loop. The performance of \acronym 4EC is comparable to the analytic model and \acronym 4EC achieves the swing-up of the physical Furuta pendulum, where system identification does not, despite having a lower nMSE compared to the analytic model. This shows that a low nMSE does not necessarily imply good control performance.

\section{Conclusion}
In this paper, we introduced the concept of Deep Lagrangian Networks for energy control (\acronym 4EC), a learning to control approach that combines the flexibility of deep learning with the insights from control theory. This combination is enabled only because \acronym 4EC imposes Lagrangian Mechanics, conservation of energy and energy coherence on a generic deep network and hence, learns a physically plausible model. We showed that \acronym is able to learn the inertial-, centripetal-, Coriolis-, gravitational- and frictional forces and the potential and kinetic energy from sensor data containing only joint configuration and motor torque. Therefore, learning these forces and energies is unsupervised and does not require any knowledge about the kinematic structure. Other model learning algorithms either require the kinematic structure to learn these components such as system identification or cannot learn the force components or system energies such as neural networks or Gaussian process regression. The qualitative and quantitative offline evaluation showed that the normalized MSE of \acronym in comparison to the analytic model is comparable for the simulated systems and better for the physical systems. For the online control task, \acronym 4EC accomplishes the swing-up of the physical Cartpole and Furuta pendulum from different starting configurations by computing the system energies within a $500$Hz real-time control loop. In contrast, the system identification model only achieves the successful swing-up of the physical Cartpole but not the physical Furuta pendulum, despite having comparable nMSE to \acronym 4EC. 

\bibliography{iclr2019_conference.bib}
\bibliographystyle{IEEEtran}

\addtolength{\textheight}{-12cm}   % This command serves to balance the column lengths

\end{document}